\documentclass[conference]{IEEEtran}
\usepackage{times}

\usepackage[numbers]{natbib}
\usepackage{multicol}
\usepackage[bookmarks=true]{hyperref}
\usepackage{soul}
\usepackage{url}
\usepackage[utf8]{inputenc}
\usepackage[small]{caption}
\usepackage{subcaption}
\usepackage{graphicx} 
\usepackage{amsmath}
\usepackage{amsthm}
\usepackage{booktabs}
\usepackage{algorithm}
\usepackage{algorithmic}

\usepackage{xcolor}

\usepackage{hhline}

\usepackage{wrapfig}
\usepackage{lipsum}
\newtheorem{definition}{Definition}

\title{Planning Multimodal Exploratory Actions for Online Robot Attribute Learning}

\author{\authorblockN{Xiaohan Zhang}
\authorblockA{State University of New York at Binghamton\\
Binghamton, NY 13902, USA\\
Email: xzhan244@binghamton.edu}
\and
\authorblockN{Jivko Sinapov}
\authorblockA{Tufts University\\
Medford, MA 02155, USA\\
Email: jivko.sinapov@tufts.edu}
\and
\authorblockN{Shiqi Zhang}
\authorblockA{State University of New York at Binghamton\\
Binghamton, NY 13902, USA\\
Email: zhangs@binghamton.edu}}

\begin{document}
\maketitle

\begin{abstract}
Robots frequently need to perceive object attributes, such as ``red,'' ``heavy,'' and ``empty,'' using multimodal exploratory actions, such as ``look,'' ``lift,'' and ``shake.''
Robot attribute learning algorithms aim to learn an observation model for each perceivable attribute given an exploratory action. 
Once the attribute models are learned, they can be used to identify attributes of new objects, answering questions, such as ``Is this object red and empty?'' 
Attribute learning and identification are being treated as two separate problems in the literature. 
In this paper, we first define a new problem called online robot attribute learning~(On-RAL), where the robot works on attribute learning and attribute identification simultaneously. 
Then we develop an algorithm called information-theoretic reward shaping~(ITRS) that actively addresses the trade-off between exploration and exploitation in On-RAL problems. 
ITRS was compared with competitive robot attribute learning baselines, and experimental results demonstrate ITRS' superiority in learning efficiency and identification accuracy. 
\footnote{Additional materials are available at \url{https://sites.google.com/view/on-ral}}
\end{abstract}

\IEEEpeerreviewmaketitle

\section{Introduction}
\label{sec:intro}

Intelligent robots are able to interact with objects through exploratory behaviors in real-world environments. 
For instance, a robot can take a $look$ behavior to figure out if an object is ``red'' using computer vision methods. 
However, vision is not sufficient to answer if an opaque bottle is ``full'' or not, and behaviors that support other sensory modalities, such as $lift$ and $shake$, become necessary. 
Given the sensing capabilities of robots and the perceivable properties of objects, it is important to develop algorithms to enable robots to use multimodal exploratory behaviors to identify object properties, answering questions such as ``\emph{Is this object red and empty?}''
In this paper, we use \textbf{attribute} to refer to a perceivable property (of an object) and use \textbf{behavior} to refer to an exploratory action that a robot can take to interact with the object. 

Robot multimodal perception is a challenge for several reasons. 
First, exploratory behaviors can be costly, and even risky in the real world. 
For instance, to $shake$ a water bottle to identify the value of attribute ``empty'', the robot must first $grasp$ and $lift$ it. 
Those behaviors take time and can break the bottle in case of failed grasps. 
Second, those behaviors are not equally useful for recognizing different attributes.
For instance, $lift$ is more useful than $look$ for ``heavy,'' while $look$ works much better for ``shiny.''
\textbf{Robot attribute learning}~(RAL) algorithms aim to learn an observation model for each attribute given an exploratory behavior and play a key role in robot multimodal perception. 
Most existing RAL algorithms are considered \textbf{offline}: the robot learns the attributes by interacting with objects \emph{without} considering data collection costs. 
In the evaluation phase, the robot uses the learned attributes to identify attributes of new objects (i.e., attribute identification). 
In this research, we are concerned with a novel \textbf{online} RAL~(On-RAL) setting, where the robot needs to learn an action policy for interacting with objects toward efficient attribute learning and accurate attribute identification at the same time. 

\begin{figure*}
\begin{center}
    \vspace{-1em}
    \includegraphics[width=0.95\textwidth]{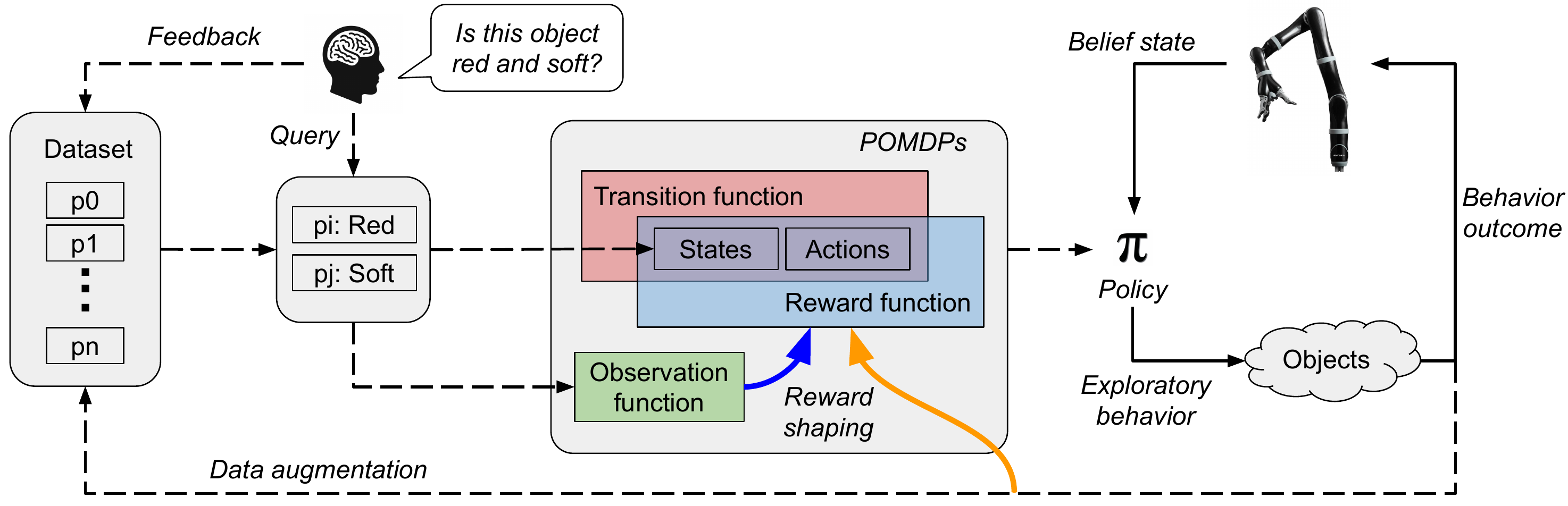}
    \vspace{-.5em}
    \caption{An overview of the ITRS algorithm. A human user will choose an object and ask a query such as ``\emph{Is this object red and soft?}''. The robot will generate a perception model on the specified attributes, i.e., ``red'' and ``soft''. Queried attributes and the corresponding perception model then will be used to construct states and the observation function of the POMDP model respectively. The reward function will be shaped by the quality of the observation function and the robot's experience. The robot uses the generated POMDP model to compute a policy $\pi$ and interacts with the queried object. Newly-perceived feature data will be used to update the robot's experience and augment the dataset. Humans will give feedback to the robot's answer and attach labels to the feature data points.}
    \label{fig:overview}
    \vspace{-1em}
\end{center}
\end{figure*}

On-RAL faces the fundamental trade-off between exploration and exploitation. 
A trivial solution is to let the robot optimize its behaviors solely on attribute identification as if the attributes have been learned already. 
In doing so, the robot still learns the observation models of attribute-action pairs as it becomes more experienced, but this trivial solution lacks a mechanism for actively improving its long-term attribute identification performance. 
The main contribution of this paper is an algorithm, called \emph{information-theoretic reward shaping} (\textbf{ITRS}), for On-RAL problems. 
ITRS, for the first time, equips a robot with the capability of optimizing its sequential action selection toward (efficiently and accurately) learning and identifying attributes at the same time, as shown in Figure~\ref{fig:overview}. 
ITRS has been evaluated using two datasets: one dataset, called \textbf{CY101}, contains 101 objects with ten exploratory behaviors and seven types of sensory modalities~\cite{tatiya2019deep}; and the other, called \textbf{ISPY32}, includes 32 objects with eight behaviors and six types of modalities~\cite{thomason2016learning}. 
Compared with existing methods from the RAL literature~\cite{amiri2018multi,thomason2018guiding}, ITRS reduces the overall cost of exploration in the long term while reaching a higher accuracy of attribute identification. 


\section{Related Work}


\subsection{Multimodal Perception in Robotics}
Significant advances have been achieved recently in computer vision, e.g.,~\cite{krizhevsky2012imagenet,redmon2016you} and natural language processing, e.g.,~\cite{devlin2018bert,brown2020language}. 
While language and vision are important communication channels for robotic perception, many object properties cannot be detected using vision alone~\cite{gibson1988exploratory} and people are not always available to verbally provide guidance in exploration tasks.
Therefore, researchers have jointly modeled language and visual information for multimodal text-vision tasks~\cite{radford2021learning}. 
However, many of the most common nouns and adjectives (e.g., ``soft'', ``empty'') have a strong non-visual component~\cite{lynott2009modality} and thus, robots would need to perceive objects using additional sensory modalities to reason about and perceive such linguistic descriptors. 
To address this problem, several lines of research have shown that incorporating a variety of sensory modalities is the key to further enhance the robotic capabilities in recognizing multisensory object properties (see~\cite{bohg2017interactive} and~\cite{li2020review} for a review). 
For example, visual and physical interaction data yields more accurate haptic classification for objects~\cite{gao2016deep}, and non-visual sensory modalities (e.g., $audio$, $haptics$) coupled with exploratory actions (e.g., $touch$ or $grasp$) have been shown useful for recognizing objects and their properties~\cite{kerzel2019neuro,gandhi2020swoosh,braud2020robot,pastor2020bayesian,Sawhney2021playing}, as well as grounding natural language descriptors that people use to refer to objects~\cite{thomason2016learning,arkin2020multimodal}. 
More recently, researchers have developed end-to-end systems to enable robots to learn to perceive the environment and perform actions at the same time~\cite{lee2019making,wang2020swingbot}.

A major limitation of these and other existing methods is that they require large amounts of object exploration data, which is much more expensive to collect as compared to vision-only datasets. A few approaches have been proposed to actively select behaviors at test time (e.g., when recognizing an object~\cite{fishel2012bayesian,sinapov2014grounding} or when deciding whether a set of attributes hold true for an object~\cite{amiri2018multi}). 
One recent work has also shown that robots can bias which behavior to perform at training time (i.e., when learning a model grounded in multiple sensory modalities and behaviors) but they did not learn an actual policy for doing so~\cite{thomason2018guiding}. 
Different from existing work, we propose a method for learning a behavior policy for object exploration that a robot can use when learning to ground the semantics of attributes. 



\subsection{Planning under Uncertainty}
Decision-theoretic methods have been developed to help agents plan behaviors and address uncertainty in non-deterministic action outcomes~\cite{puterman2014markov,sutton2018reinforcement}. 
Existing planning models such as partially observable Markov decision process~(POMDP)~\cite{kaelbling1998planning}, belief space planning~\cite{platt2010belief} and Bayesian approaches~\cite{ross2011bayesian} have shown great advantages for planning robot perception behaviors, because robots need to use exploratory actions to estimate the current world state.
To learn semantic attributes, robots frequently need to choose multiple actions, so POMDP which is useful for long-term planning is particularly suitable.
Many of the POMDP-based robot perception methods are vision-based~\cite{sridharan2010planning,eidenberger2010active,zheng2020multi,zhang2013active}. 
Compared to those methods, our robot takes advantage of non-visual sensory modalities, such as $audio$ and $haptics$. 

Work closest to this research plans under uncertainty to interact with objects using multimodal exploratory actions~\cite{amiri2018multi}, where they modeled the mixed observability~\cite{ong2010planning} in domains of a robot interacting with objects (we use their work as a baseline approach in experiments). 
The work of~\citet{amiri2018multi} and this work share the same spirit from the planning and perception perspectives. 
The main difference is that their work assumed that sufficient training data and annotations are available for the robot to learn the perception models of its exploratory actions.
In comparison, we consider a more challenging setting, called ``Online RAL,'' where the data collection and task completion processes are simultaneous.

\subsection{Robot Attribute Learning (RAL)}
To select actions to identify objects' perceivable properties (e.g., ``heavy,'' ``red,'' ``full,'' and ``shiny''), robots need observation models for their exploratory actions. 
Researchers have developed algorithms to help robots determine the presence of possibly new attributes~\cite{konidaris2018skills} and learn observation models of objects' perceivable properties (i.e., attributes) given different exploratory actions~\cite{thomason2016learning,thomason2017opportunistic,sinapov2016learning}. 
In the case where the object attributes refer to the object's function, they are then referred to as 0-order affordances \cite{aldoma2012supervised}. 
Those methods focused on learning to improve the robots' perception capabilities. 
Once the learning process is complete, a robot can use the learned attributes to perform tasks, such as attribute identification (e.g., to tell if a bottle is ``heavy'' and ``red''). 
Compared to those learning methods, we consider an online multimodal RAL setting, where the robot learns the attributes (an exploration process) and uses the learned attributes to identify object properties (an exploitation process) at the same time. 
The exploration-exploitation trade-off is a fundamental decision-making challenge in unknown environments. 
While the problem has been studied in multi-armed bandit~\cite{katehakis1987multi} and reinforcement learning settings~\cite{sutton2018reinforcement}, it has not been studied in RAL contexts.

\begin{figure}
\begin{center}
    \vspace{.5em}
    \includegraphics[height=7.3em]{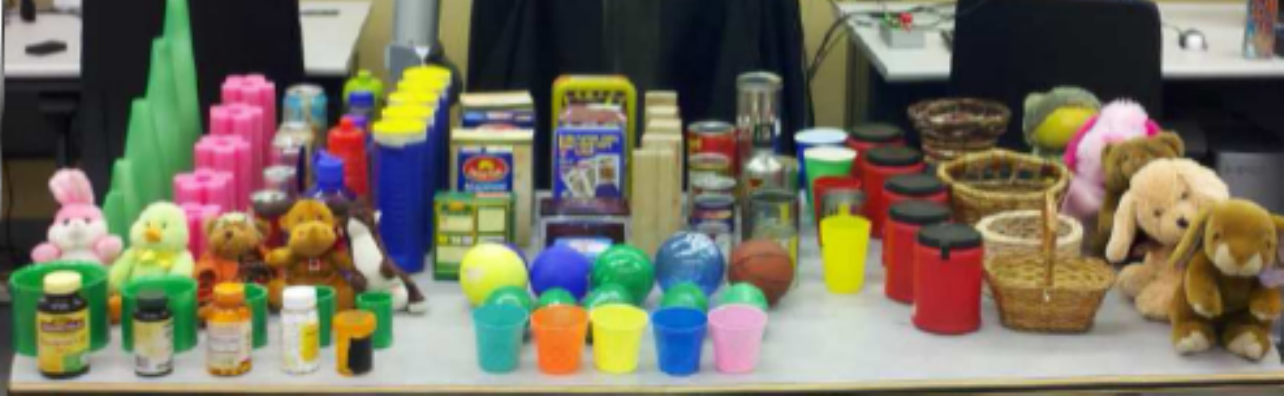}
    \caption{Everyday objects used in the demonstrations and evaluations of this research~\protect\cite{tatiya2019deep}. }
    \vspace{-1.5em}
    \label{fig:objects}
\end{center}
\end{figure}

\section{Theoretical Framework}
In this section, we first formally define three types of robot multimodal perception problems, including On-RAL. 
We then describe information-theoretic reward shaping (ITRS), a novel algorithm for On-RAL problems. 

\subsection{Problem Definitions}
\label{sec:problem}
A robot has a set of actions, such as $look$, $push$, and $lift$, that can be used for interacting with everyday objects as shown in Figure~\ref{fig:objects}. 
Let $o \in Obj$ be an object and $a \in A^e$ be an \emph{exploratory action}. 
Each action is coupled with a set of sensory modalities, e.g., $vision$, $haptics$, and $audio$. 
We use $m \in \mathcal{M}$ to refer to a sensory modality. 
This action-modality coupling is formulated using function $\Gamma$: 
\begin{align}
  \mathcal{M}_a = \Gamma(a)
\end{align}
where $\mathcal{M}_a \subseteq \mathcal{M}$. 
For example, $\{audio,haptics,vision\}=\Gamma(push)$ means that action $push$ produces signals from $audio$, $haptics$, and $vision$ modalities. 

Each action-modality pair specifies a set of \emph{viable} combinations $\mathcal{C}$, and $c\in \mathcal{C}$ is called a sensorimotor context: 
\begin{align}
  \mathcal{C} = A^e ~\otimes^{\Gamma} \mathcal{M}
  \label{eqn:c}
\end{align}
where $\otimes^{\Gamma}$ is a \emph{viable Cartesian product} operation that outputs only those viable pairs from $A^e \times \mathcal{M}$, and the viability is determined by $\Gamma$. 
We use $c^m_a$ to refer to the context specified by $m$ and $a$. 
For instance, $(look,vision)$ is a viable context, whereas $(look, audio)$ is not. 
When $a$ is performed on object $o$, for each $m \in \mathcal{M}_a$, the robot is able to record a data instance $f^m_a$. 
We use $f_a$ to represent the instances of all modalities that a robot receives after performing $a$. 

Let $\mathcal{P}$ specify a set of attributes that are used to describe objects in a domain. 
Given object $o$, $v^p$ is either \emph{true} or \emph{false}, depending on if $p$ applies to $o$, where we use $ID(p,o)$ to refer to this attribute identification function. 
Here we ``override'' function $ID$ to use it to process a set of attributes:
\begin{align}
  \textbf{v}=ID(\textbf{p},o)    
\end{align}
where $\textbf{v}=[v^{p_0}, v^{p_1},\cdots]$, $\textbf{p}=[p_0, p_1, \cdots]$, and $v^{p_i}$ is the value of attribute $p_i$ of object $o$. 
For instance, given a red empty object (i.e., $o$) and two attributes of $blue$ and $empty$ (i.e., $\textbf{p}$), the $ID(\textbf{p}, o)$ outputs $[false, true]$.

\begin{definition}[\textbf{Off-RAL}]
At training time, the input includes a set of labeled sensory data instances, each in the form of $(f_a, p):v^p$.
Solving an \textbf{offline RAL} (Off-RAL) problem produces a binary classifier:
$$
    \Psi(f_a, p), \textnormal{~for each pair of~} a\in A^e \textnormal{~and~} p\in \mathcal{P}
$$ 
\end{definition}

At testing time, given object $o$, a robot collects data instances $f_a$ after performing action $a$ and $\Psi(f_a, p)$ outputs \emph{true} or \emph{false} estimating if attribute $p$ applies to $o$. 

\begin{definition}[\textbf{RAI}]
Solving a \textbf{robot attribute identification} (RAI) problem produces policy $\pi$ that sequentially selects action $a\in A^e$ to identify the value of: 
$$
  ID(\textbf{p},o), \textnormal{  given  }  \Psi
$$
where the objective is to identify the attribute(s) in each identification task while minimizing the total action cost. 
\end{definition}

\begin{definition}[\textbf{On-RAL}]
Solving an \textbf{online RAL} (On-RAL) problem produces policy $\pi$ that sequentially selects action $a\in A^e$ to identify the value of:  
$$
  ID(\textbf{p},o)
$$

At execution time, after performing $a$ to identify $p$, the robot collects data in the form of $(f_a, p)$. 
After each identification task, the robot receives $\textbf{v}$, the values of attributes $\textbf{p}$. 
The objective is to minimize the discounted cumulative action cost and maximize the success rate of attribute identification. 
\end{definition}

\noindent
{\bf Remarks: }Rational RAI agents treat individual attribute identification tasks independently, whereas
rational On-RAL agents learn from the data collected in early tasks, trading off early-phase performance for long-term performance. 
Although existing RAI methods used different action policies, they all require an Off-RAL algorithm (for computing $\Psi$) running as a \emph{preprocessing} step~\cite{sinapov2016learning,amiri2018multi,thomason2018guiding}. 
For instance, \citet{thomason2018guiding} used a random strategy, and \citet{amiri2018multi} used a planning under uncertainty approach.

\begin{figure}
\begin{center}
    \includegraphics[height = 6.2cm]{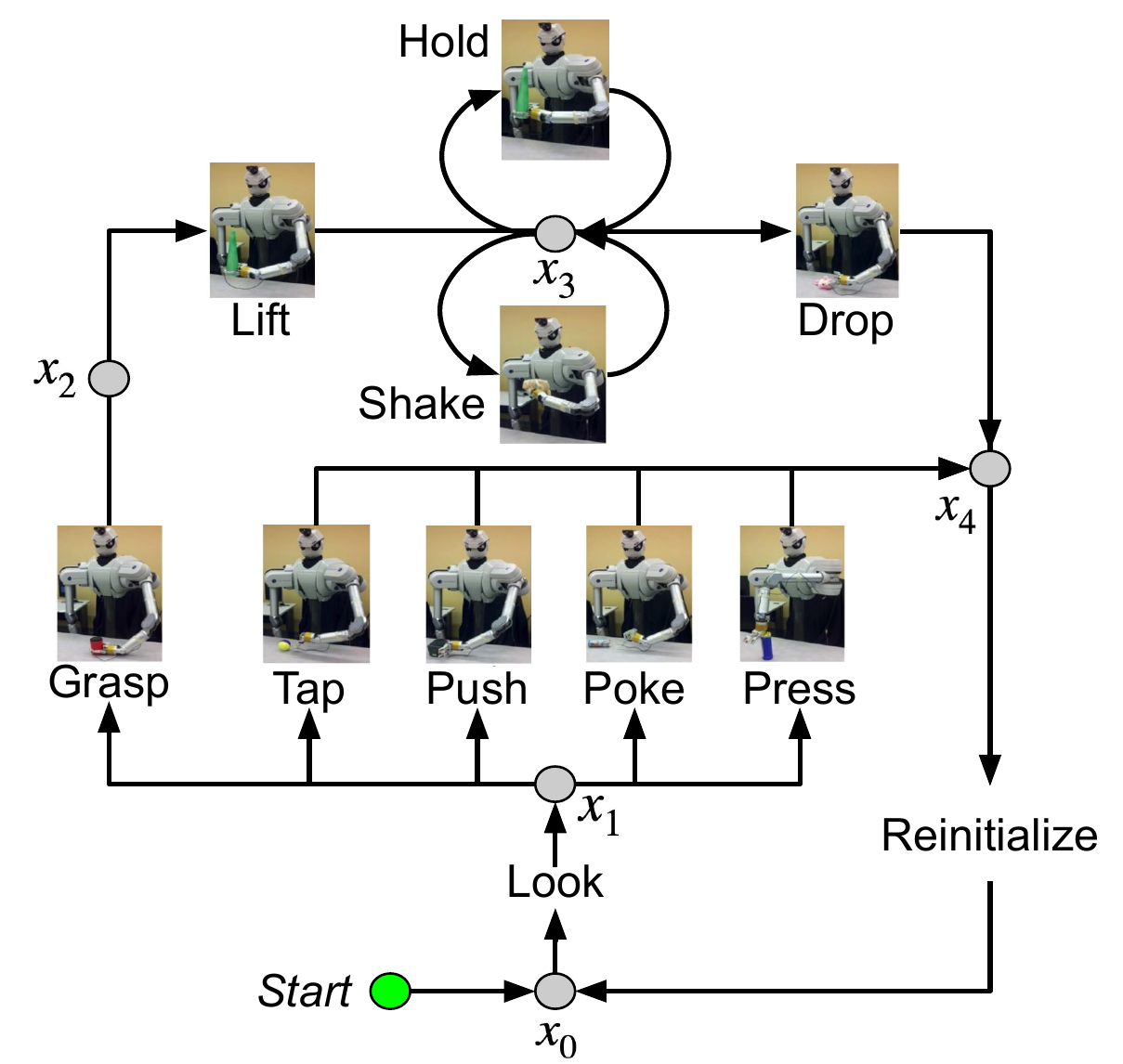}
    \vspace{-.5em}
    \caption{Transition diagram among the $\mathcal{X}$ states led by exploratory actions $A^e$.}
    \vspace{-1.5em}
    \label{fig:trans}
\end{center}
\end{figure}

\subsection{Dynamically Constructed POMDPs}
\label{app:pomdp}
Aiming at computing an action policy toward solving On-RAL problems, we dynamically construct partially observable MDPs~(POMDPs)~\cite{kaelbling1998planning}. 
POMDPs can be defined in the form of $(S,\ A,\ T,\ R,\ Z,\ O,\ \gamma)$, where $S$, $A$, and $Z$ are the sets of states, actions, and observations respectively; $T$, $R$, and $O$ are functions of transition, reward, and observation respectively. 
$\gamma$ is a discount factor. 


The state space $S$ of the POMDP is a Cartesian product of two components, $\mathcal{X}$ and $\mathcal{Y}$:
$$
  S = {(x,y) | x \in \mathcal{X}, y \in \mathcal{Y}}
$$
where $\mathcal{X}$ is the state set specified by fully observable domain variables. 
In our case, $\mathcal{X}$ includes a set of five states {$x_0, ..., x_4$}, as shown in Figure~\ref{fig:trans}, and a terminal state $term \in \mathcal{X}$ that identifies the end of an episode.
The states in $\mathcal{X}$ are fully observable, meaning that the robot knows the current state of the robot-object system, e.g., whether $grasp$ and $drop$ are successful or not. 
$\mathcal{Y}$ is the state set specified by partially observable domain variables ($N$ queried attributes), {$p_0, p_1, ... , p_{N-1}$}. 
Thus, $|\mathcal{Y}| = 2^N$. 
For instance, given an object description that includes three attributes (e.g., a ``red'' ``empty'' ``container''), $\mathcal{Y}$ includes $2^3$ states. 

Let $A: A^e\cup A^r$ be the action set that is available to the robot. 
$A^e$ includes the object \emph{exploration} actions (Figure~\ref{fig:trans}), and $A^r$ includes a set of \emph{reporting} actions that deterministically lead the state transition to \textit{term} and are used for attribute identification. 
Continuing the ``red empty container'' example, there are three binary variables and $|A^r| = 8$, each $A^r$ corresponding to $y \in \mathcal{Y}$.



Let $T_{\mathcal{X}}: \mathcal{X}\times A\times \mathcal{X}\rightarrow [0,1]$ be the state transition function in the fully observable component of the current state. 
Most exploration actions may be unreliable to some degree. 
For instance, $p(x_3,~\textit{drop},~x_4)=0.95$ in our case, indicating there is a small probability the object is stuck in the robot's hand. 
$T_{\mathcal{Y}}:\mathcal{Y}\times A\times \mathcal{Y}\rightarrow [0,1]$ is an identity matrix, because object attributes do not change over time.

Let $Z$ be a set of observations and let the observation function $O(s, a, z)$ specify the probability of observing $z \in Z$ after taking action $a$ in state $s$.
We calculate the probability using confusion matrix $\Theta_p^a \in R^{2 \times 2}$:
\begin{align}
O(s, a, z) = \mathbf{Pr}(\mathbf{p}^z | \mathbf{p}^s, a) = \prod_{i=0}^{N-1} \Theta_{p_i}^a(p_i^s, p_i^z)
\label{observation function}
\end{align}
where $\Theta_p^a \in R^{2 \times 2}$ is a confusion matrix for attribute $p$ and action $a$; 
$\mathbf{p}^s$ and  $\mathbf{p}^z$ are the vectors of true and observed values (0 or 1) of the attributes; 
$p_i^s$ (or $p_i^z$) is the true (or observed) value of the $i^{th}$ attribute; and 
$N$ is the total number of attributes in the query.
The transition system and the computation of $\Theta_p^a \in R^{2 \times 2}$ are adapted from those of~\cite{amiri2018multi}. 
To alleviate the scalability issue of POMDPs, we use a strategy of dynamically constructing minimal POMDPs to model only those attributes necessary to the current query~\cite{zhang2017dynamically}.

\subsection{Algorithm Description}
\label{sec:reward}

We first introduce our information-theoretic reward function and then present a novel algorithm for On-RAL problems. 

In On-RAL problems, the robot needs to optimize its behaviors toward not only improving the accuracy of attribute identification but also minimizing the cost of exploratory actions. 
We introduce the two factors of \emph{perception quality} and \emph{interaction experience} into the reward design of POMDPs to achieve the trade-off between exploration (actively collecting data for attribute learning) and exploitation (using the learned attributes for identification tasks). 

Let $Ent(s, a)$ be the entropy of the distribution over $Z$, given $s$ and $a$, and is used for indicating the \textbf{perception quality} of exploratory action $a$ over the $y$ component of $s$:
\begin{align}
 Ent(s, a) = - \sum_{z_i \in Z}O(z_i|s,a)\log_2 O(z_i|s,a)
\label{entropy}
\end{align}
where $z_i$ is the $i^{th}$ observation and $O(z_i|s,a)$ is the probability of observing $z_i$ in state $s$ after taking action $a$. $O(z_i|s,a)$ is computed using the data instances gathered in the On-RAL process -- see Definition 3.


Let $IE(p, a)$ be the \textbf{interaction experience} of applying action $a$ to identify attribute $p$, and is in the form of: 
\begin{align}
    IE(p,a) = \frac{1}{\delta} \cdot | \{ f \in \textbf{f}_a^m, labeled(f,p)=true \}|
\label {eqn:experience}
\end{align}
where $\textbf{f}_a^m$ is a set of instances in context $c^m_a$ that a robot has collected so far, and $labeled(f,p)$ returns \emph{true} if $f$ has been labeled w.r.t. $p$, where the value $v^p$ is \emph{true} or \emph{false}. 
$\delta$ is a sufficiently large integer to ensure $IE(p,a)$ is in the range of [0,1). 
A lower value of $IE(p,a)$ reflects a higher need of further exploring $(p,a)$.

\noindent
{\bf Information-theoretic Reward:}
We use $R^{real}(s,a,s')$ to refer to the standard reward function that rewards (or penalizes) successful (or unsuccessful) attribute identifications~\cite{amiri2018multi}. 
Building on the concepts of perception quality and interaction experience, our information-theoretic reward function is defined as follows: 
\begin{align}
R(s, a, s') = & R^{real}(s,a,s') + \nonumber \\
  & \alpha \cdot Ent(s, a) - \beta \cdot IE(p, a)  
\label{reward}
\end{align}
where $\alpha$ and $\beta$ are natural numbers and used for adjusting how much perception quality and interaction experience are considered in reward shaping. 
Informally, when $O(z|s, a)$ is close to being uniform, the perception model of $(s, a)$ is poor, and the value of $Ent(s, a)$ is high. 
As a result, our new reward function will encourage the robot to take action $a$ by offering extra reward $\alpha \cdot Ent(s, a)$. 
When the robot is experienced in applying $a$ to identify attribute $p$, $IE(p, a)$ will be high. 
In this case, an extra penalty of $\beta \cdot IE(p, a)$ will discourage the robot from taking those well-explored actions. 
In comparison to standard POMDPs, where reward and observation functions are independently developed, ITRS enables the reward function to adapt to the changes of the observation function. 

\vspace{.5em}

\noindent
{\bf ITRS Algorithm:}
Algorithm~\ref{alg:ITRS} presents ITRS for active On-RAL problems. 
The inputs of ITRS include attribute set $\mathcal{P}$, transition function $T_{\mathcal{X}}$, action set $A^e$, POMDP solver $Sol$, parameters $\alpha$ and $\beta$, naive reward function $R^{real}(s,a,s')$, and dataset $\mathcal{D}^{pre}$ for pretraining. 
ITRS does not have a termination condition. 

ITRS starts with initializing the interaction experience function with zeros for all $(p, a)$ pairs, and then initializes dataset $\mathcal{D}$ that will be later augmented as the robot interacts with objects (Lines~\ref{l:initIE} and~\ref{l:initD}). 
In each iteration of the main loop (Lines~\ref{trial loop start}-\ref{trial loop end}), ITRS takes an identification query from people (Line~\ref{query}), constructs a POMDP model (Lines~\ref{construct pomdp start}-\ref{construct pomdp end}), computes its policy, uses the policy to interact with an object (Lines~\ref{pomdp start}-\ref{pomdp end}), and augments its dataset for improving the POMDP model in the next iteration (Lines~\ref{update start}-\ref{update end}). 

In the first inner loop (Lines~\ref{pomdp start}-\ref{pomdp end}), the robot interacts with an object based on the generated policy. 
$\pi$ suggests an action at each state $b$.
The robot then executes the action and makes an observation.
Based on the action and observation, the robot updates its belief using the Bayesian rule. 
After selecting each action, ITRS records this action along with its collected feature data (Lines~14 and 15).
In Line 19, we ask people to provide the label $y$ for the collected data of $\{f^{c_0}, f^{c_1}, \cdots\}$.
The final step is to iterate over all selected actions to update function $\Lambda$, augment $\mathcal{D}$, and calculate the new interaction experience (Lines~\ref{update start}-\ref{update end}).

\begin{algorithm}[t!]\small
\caption{ITRS algorithm}\label{alg:ITRS}
\begin{algorithmic}[1]
\REQUIRE $\mathcal{P}$; $T_{\mathcal{X}}$; $A^e$; $Sol$; $\alpha$; $\beta$; $R^{real}(s,a,s')$; $\mathcal{D}^{pre}$
\STATE {Initialize $IE(p,a) = 0$ for each action $a \in A$ and $p \in \mathcal{P}$} \label{l:initIE}
\STATE {Let online training dataset $\mathcal{D} = \mathcal{D}^{pre}$} \label{l:initD}
\REPEAT \label{trial loop start}
\STATE {Take queried attribute(s) $\textbf{p}$ from human, where $\textbf{p} \subseteq \mathcal{P}$} \label{query}
\STATE {Generate $S$, $T_{\mathcal{Y}}$ and $Z$ using $\textbf{p}$} \label{construct pomdp start}
\STATE {Compute confusion matrix $\Theta_{p}^{a}$ using $\mathcal{D}$ where $p \! \in \! \textbf{p}$, $a \! \in \! A$}
\STATE {Generate $O(z|s,a)$ with $\Theta_{p}^{a}$ for $p \in \textbf{p}$ using Eqn.~\eqref{observation function}}
\STATE {Compute $Ent(s,a)$ using Eqn.~\eqref{entropy}}
\STATE {Generate $R(s,a,s')$ with $R^{real}(s,a,s')$ using Eqn.~\eqref{reward}}
\STATE {Compute policy $\pi$ using $Sol$ for $(S,\ A,\ T,\ R,\ Z,\ O,\ \gamma)$} \label{construct pomdp end}
\STATE {Initialize action set $A^{select}$ and feature set $\mathcal{F}$ with $\emptyset$}
\STATE {Uniformly initialize belief $b$}
\WHILE {Current state $s$ is not $term$} \label{pomdp start}
\STATE {Select action $a$ with $\pi$ based on $b$, append $a$ to $A^{select}$, and execute $a$} \label{l:policy}
\STATE {Record data instances $f_a$, and $\mathcal{F} \leftarrow \mathcal{F} \cup \{f_a\}$}
\STATE {Make an observation $z$ where $z \in Z$}
\STATE {Update $b$ with $z$ and $a$ using Bayesian rule}
\ENDWHILE \label{pomdp end}
\STATE {Ask human to provide $v^p$ for each $p \in \textbf{p}$ as label(s) for $\mathcal{F}$ }
\FOR {each $a$ in $A^{select}$} \label{update start}
\STATE {Update $\mathcal{D}$ using $\mathcal{F}$ and $v^p$ for each $p \in \textbf{p}$}
\STATE {Update $IE(p, a)$ for $a \in A$ and $p \in \textbf{p}$ using Eqn.~\eqref{eqn:experience}}
\ENDFOR \label{update end}
\UNTIL {end of interactions} \label{trial loop end}

\end{algorithmic}
\end{algorithm}

Intuitively, we aim to encourage the robot to select exploratory action $a\in A^e$ from those actions, where the perception model of $(s, a)$ is of poor quality, and there is relatively limited experience of applying $a$ to attribute $p$, i.e., the experience of $(p, a)$ is limited. 





\section{Experimental Evaluation}
\label{sec:exp}

The key hypothesis is that ITRS outperforms existing RAL algorithms in learning efficiency and task completion accuracy (there does not exist an On-RAL algorithm in the literature). 
The first baseline is referred to as ``\textbf{Random Legal},'' which corresponds to the RAL approach of~\cite{thomason2018guiding} (we did not use their linguistic component). 
It first used a supervised learning approach to compute the perception models of all attribute-action pairs (i.e., solving an Off-RAL problem) \cite{sinapov2014learning}. 
The robot considers only the ``legal'' actions (e.g., $lift$ is legal only after a successful $grasp$ action), and then randomly selects one from the legal actions. 
With an exploration budget for each trial (50 seconds and 80 seconds for one-attribute and two-attribute trials respectively), the robot is forced to report $y\in \mathcal{Y}$ of the highest belief.

The second baseline is referred to as ``\textbf{Repeated Assembly},'' which first solves an Off-RAL problem to compute $\Psi$ using all data collected so far, and then solves an RAI problem to compute policy $\pi$ for object exploration. 
This process is repeated after each batch. 
Repeated Assembly enables On-RAL behaviors by repeatedly assembling Off-RAL and RAI methods. 
Each iteration of Repeated Assembly corresponds to the work of~\cite{amiri2018multi}.

\begin{figure*}[t]
\vspace{-1.5em}
     \centering
     \begin{subfigure}[b]{0.43\textwidth}
         \centering
         \includegraphics[width=\textwidth]{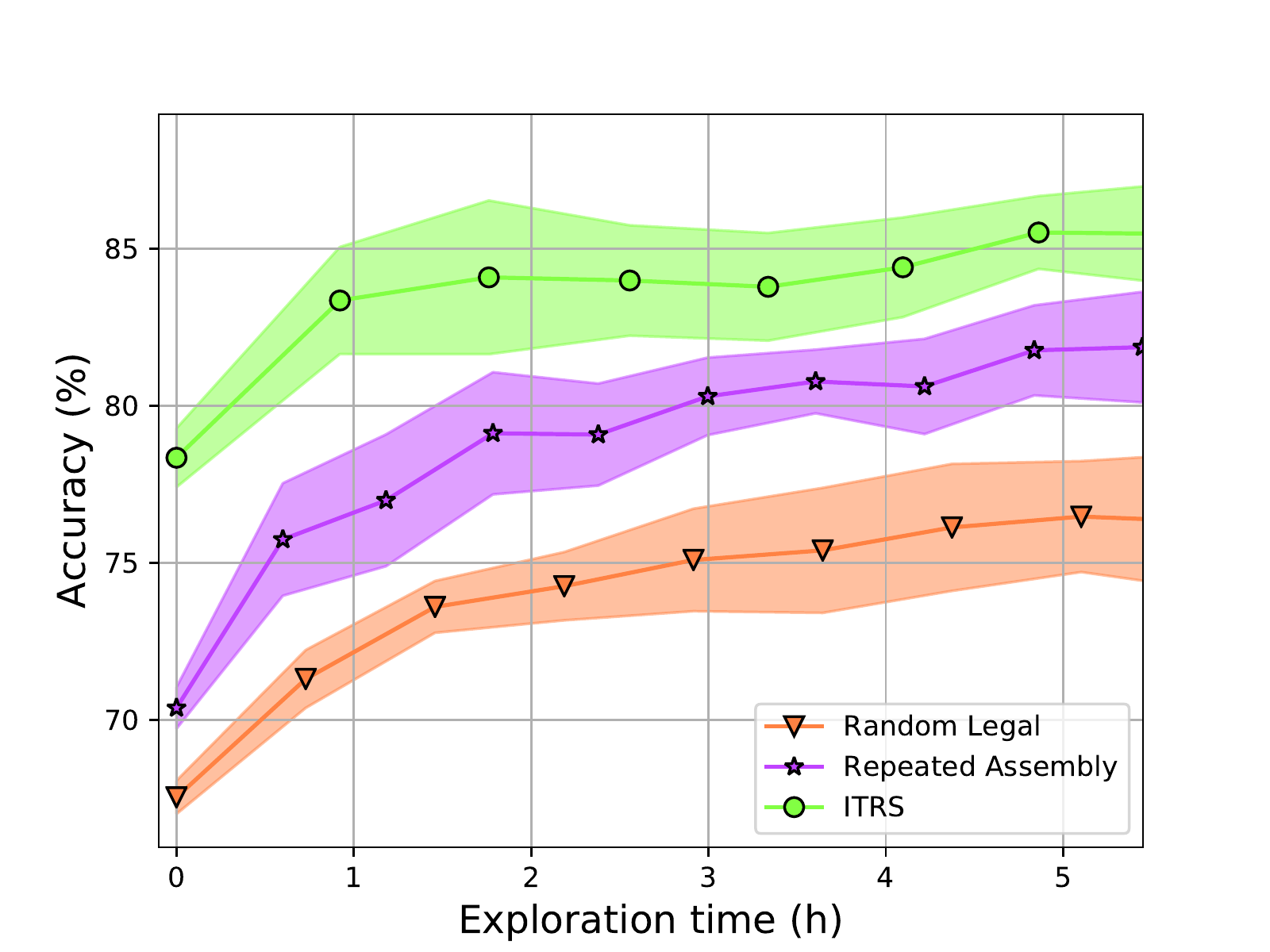}
         \vspace{-1.5em}
         \caption{One-attribute identification on CY101 dataset. }
         \label{fig:1pred}
     \end{subfigure}
     \begin{subfigure}[b]{0.43\textwidth}
         \centering
         \includegraphics[width=\textwidth]{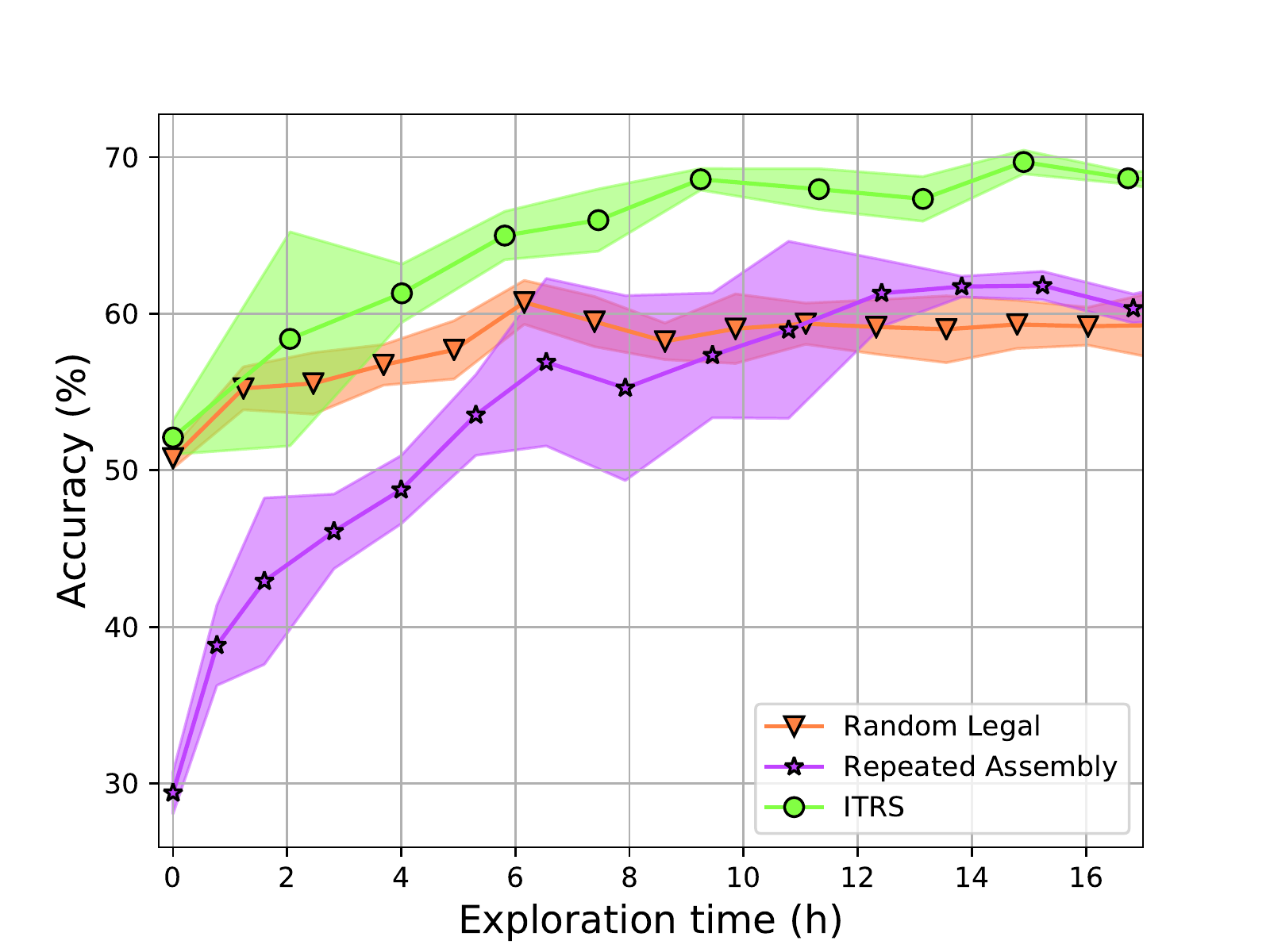}
         \vspace{-1.5em}
         \caption{Two-attribute identification on CY101 dataset. }
         \label{fig:2pred}
     \end{subfigure}
     \begin{subfigure}[b]{0.43\textwidth}
         \centering
         \includegraphics[width=\textwidth]{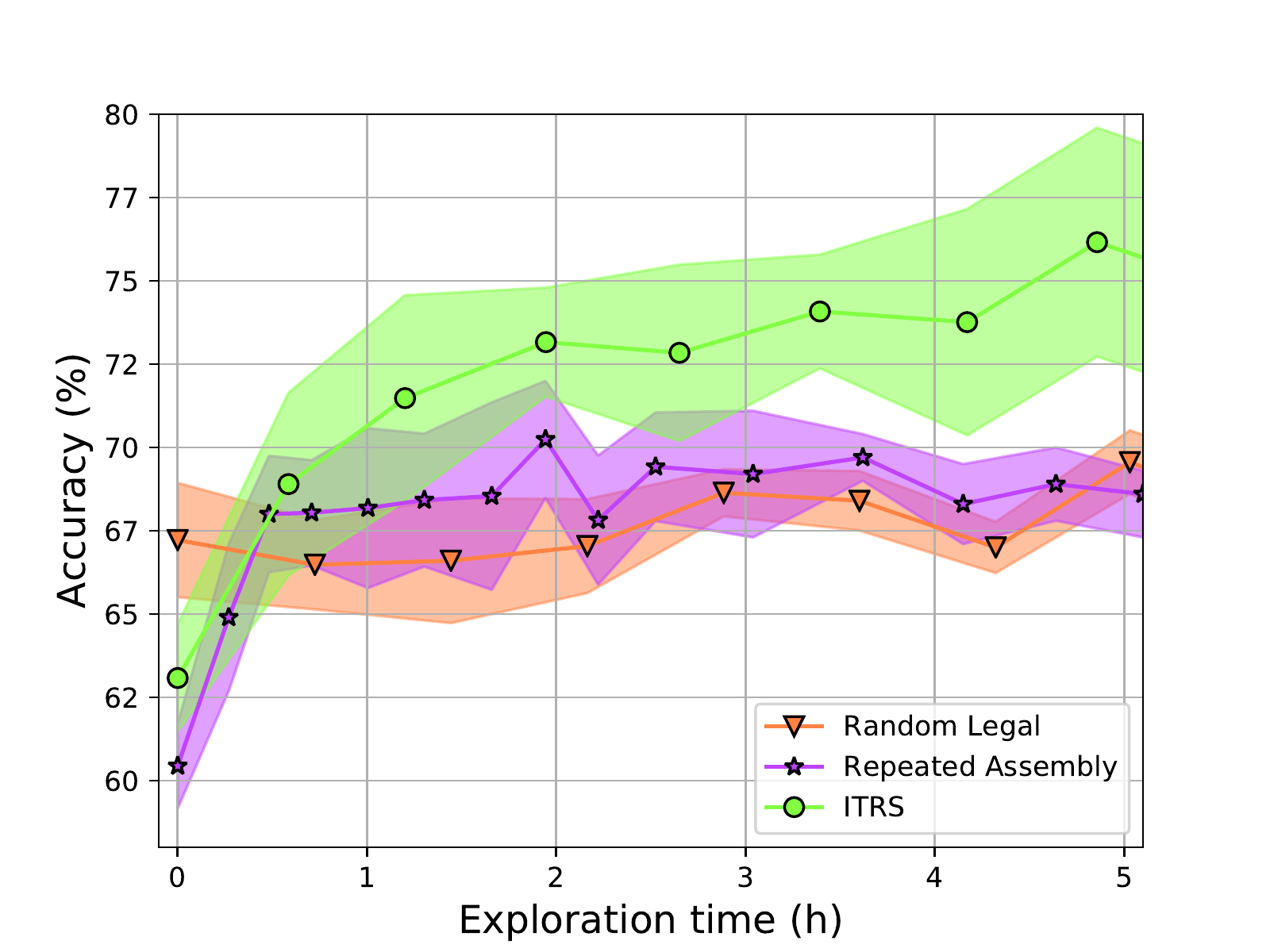}
         \vspace{-1.5em}
         \caption{One-attribute identification on ISPY32 dataset. }
         \label{fig:1pred_ijcai}
     \end{subfigure}
     \begin{subfigure}[b]{0.43\textwidth}
         \centering
         \includegraphics[width=\textwidth]{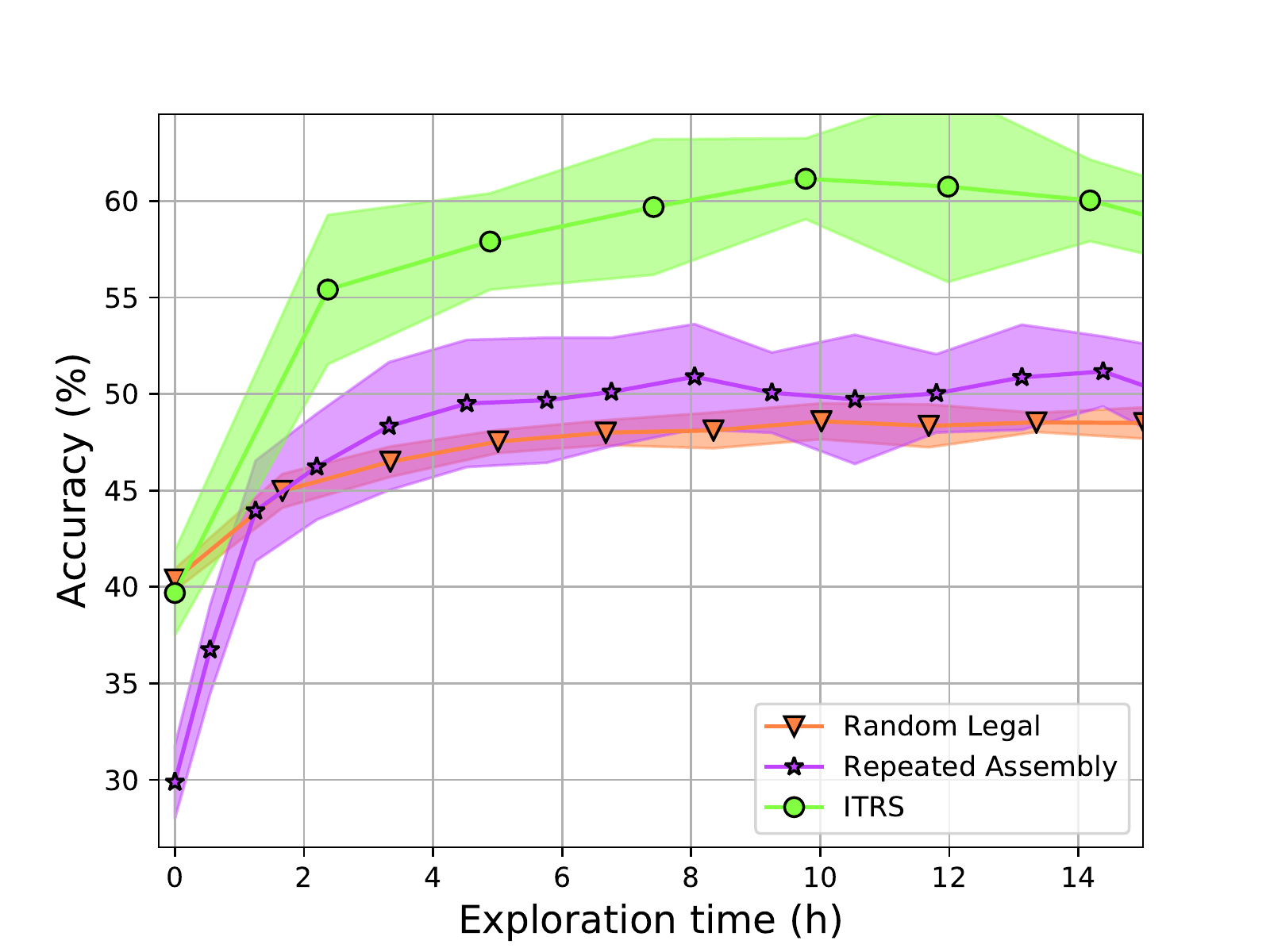}
         \vspace{-1.5em}
         \caption{Two-attribute identification on ISPY32 dataset. }
         \label{fig:2pred_ijcai}
     \end{subfigure}
     \caption{Time length of conducting exploratory actions in hours, and identification accuracy of On-RAL tasks, where we compared ITRS (ours) to two baseline strategies including \emph{Random Legal}, and \emph{Repeated Assembly}. }
     \vspace{-1em}
     \label{fig:pred}
\end{figure*}

\subsection{Experiment Setup}

\noindent
{\bf Dataset Description:}
\label{app:dataset}
Two public datasets of \textbf{CY101}~\cite{tatiya2019deep} and \textbf{ISPY32}~\cite{thomason2016learning} are used in our experiments where \textbf{CY101} (an updated version of dataset~\cite{sinapov2014grounding}) contains many more household objects and attributes.
In the \textbf{CY101} dataset, an uppertorso humanoid robot with 7-DOF arm explored 101 objects (Figure~\ref{fig:objects}) belonging to 20 different categories using 10 exploratory behaviors: $look$, $grasp$, $lift$, $hold$, $shake$, $drop$, $push$, $tap$, $poke$, and $press$ (Figure~\ref{fig:trans}).
For \textbf{ISPY32}, a robot from the Building-Wide Intelligence project~\cite{khandelwal2017bwibots} explored 32 objects using 8 exploratory behaviors: $look$, $grasp$, $lift$, $hold$, $lower$, $drop$, $push$, and $press$.
Each behavior was performed 5 times on each object in both datasets.

In order to select attributes that are learnable given the robot's exploratory behaviors, evaluations of all attributes in the two datasets were performed prior to the experiments. 
We set $|\mathcal{P}|$ to 10 and picked the attributes that are most learnable and that have enough positive and negative examples for training.
Seven different types of features including auditory, vibrotactile, finger, color, optical flow, SURF, and haptics (i.e., joint forces) were considered in \textbf{CY101}; and features of VGG, color, SURF, auditory, finger, and haptics were recorded in \textbf{ISPY32}.
For the $look$ behavior in dataset \textbf{ISPY32}, color (512-dimensional color histogram in RGB space), SURF features were extracted from the image of the object, and VGG was also extracted. 
For the other behaviors, audio, vibrotactile, flow, SURF, and haptic features were recorded via the interaction with the object.
Additionally, for the $grasp$ behavior, finger position features were recorded in the meantime.


In both datasets, we randomly split the objects into three subsets of equal sizes.
The subsets are used for pretraining (${Obj}^{pre}$), training (${Obj}^{train}$), and testing (${Obj}^{test}$) respectively. 
In the pretraining phase, the robot started with a handcrafted policy where each behavior is forced to be applied on the queried object once.
We collected feature instances with labels from those interactions and built a pretraining dataset $\mathcal{D}^{pre}$ that represents the robot's prior knowledge.

\noindent
{\bf Action Costs and Action Failures: }
Each exploratory action $a$ has a cost (planning and execution) in the range of [0.5, 22.0] that came with the datasets, and is modeled in $R^{real}(s,a,s')$. 
For instance, the cost of action $press$ (22.0) is much higher than the cost of action $look$ (0.5). 
Additionally, the reward of the reporting action was +300.0 (or -300.0) when the robot's attribute identification is correct (or incorrect).
Most actions are considered unreliable to some degree in our POMDP model and we uniformly set the failure probability to 0.05 which we did not refine in the On-RAL process.
For instance, an unsuccessful $drop$ action models the situation that the object is stuck in the robot's hand. 
We used an off-the-shelf system for solving POMDPs~\cite{kurniawati2008sarsop}.
\begin{figure*}

    \centering

    \includegraphics[width=\textwidth]{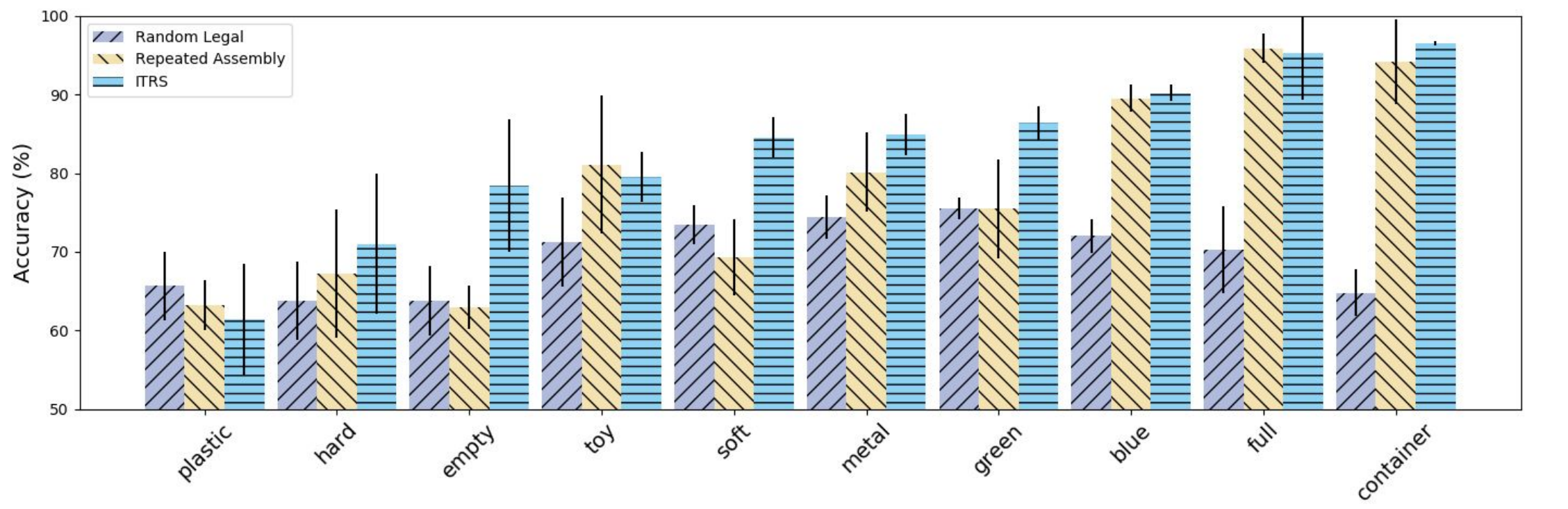}
    \vspace{-2em}
    \caption{Accuracy of attribute identification tasks. The attributes (x-axis) are ranked based on ITRS'  performance. ITRS performed the best on seven out of the ten attributes. }
    \vspace{-1em}
    \label{fig:ind}

\end{figure*}

\noindent
{\bf Queries: } 
At the beginning of each trial, an $N$-attribute identification task is assigned, i.e., $|\textbf{p}| \!\! = \!\! N$. 
In our case, $N$ was either 1 or 2. 
At the end of each trial, the robot is told if the identification was correct. 
In the case of $N=1$, the robot could learn the attribute's ground-truth value from the human's feedback. 
In the case of $N=2$, the robot could do so, only if the 2D identification was correct.


\noindent
{\bf Batch-based Learning: }
Each batch contained 40 trials in one-attribute identification tasks, and 50 trials in two-attribute tasks, where we used objects from ${Obj}^{train}$. 
After each batch, we recomputed a POMDP model, including the shaped reward function, and computed an action policy for the next batch. 
The generated POMDP policy from each batch was then evaluated using objects from ${Obj}^{test}$ in 1000 trials. 




\subsection{Illustrative Trial}

From the robot's many trials of the learning experience, we selected two trials ($T_1$ and $T_2$), where the robot faced the same object (a Coke can that has attributes ``metal,'' ``empty,'' and ``container'') and needed to answer the same question ``\emph{Is this object soft?}'' 
From the dataset, we know that the correct answer should be ``no'' (the robot did not know it). 
$T_1$ appeared at the second batch of training, and $T_2$ appeared at the ninth. 
We present both trials and explain how the robot performed better in $T_2$.


In $T_1$ (early learning phase), the robot first performed the $look$ action. 
Then, the robot had the following options: $grasp$, $tap$, $push$, $poke$ and $press$ according to Figure~\ref{fig:trans}.
Specifically, for $press$, the confusion matrix $\Theta_{soft}^{press}$ (shown in Table~\ref{tab:cm}) was nearly uniform, which is typical in the early learning phase. 
Among those ``less useful'' actions, the robot chose $grasp$.
The distribution over $\mathcal{Y}$ was changed from [0.37, 0.63] to [0.46, 0.54], where the entries represent ``not soft'' and ``soft'' respectively.
After $press$, ITRS sequentially suggested $grasp$, $lift$, $hold$ and $hold$. 
Finally, the robot reported ``positive'' that resulted in a failed trial with a total cost of 55.5 seconds.



In $T_2$ (late learning phase), 
Action $press$ became more useful for identifying attribute ``soft'' compared to $T_1$, as shown in Table~\ref{tab:cm}.
For $grasp$, $IE(soft, grasp) = 0.67$ and $\Theta_{soft}^{grasp}$ was [0.66, 0,33, 0.61, 0.38] (TN, FN, FP, TP), which meant that the robot was experienced with action $grasp$ and considered $grasp$ was not as useful as $press$. 
Accordingly, ITRS suggested $press$ instead of $grasp$ after taking $look$.
The belief over $\mathcal{Y}$ changed from [0.57, 0.43] to [0.67, 0.33]. 
After only $look$ and $press$, the robot was able to quickly report ``negative'', resulting in a successful trial with a total cost of 22.5 seconds.

From the above two trials (same query and object in different learning phases), we see how the improved perception model of $(press, soft)$ helped the robot correctly identify ``soft'' with a low cost.

\vspace{-1em}
\begin{table}[t]
\vspace{.2em}
\centering
\caption{Early and late observation models for action $press$}
\label{tab:cm}
\vspace{-.5em}
\resizebox{0.45\textwidth}{!}{%
\begin{tabular}{|c||c|c||c|c|} 
\hline
                       & \multicolumn{2}{c||}{Early phase}   & \multicolumn{2}{c|}{Late phase}      \\ 
\hhline{|=::==::==|}
\begin{tabular}[c]{@{}c@{}}Not soft\\ (Ground Truth)\end{tabular}  & 0.68               & 0.32           & 0.82               & 0.17            \\ 
\hline
\begin{tabular}[c]{@{}c@{}}Soft\\ (Ground Truth)\end{tabular}     & 0.50               & 0.50           & 0.20               & 0.80            \\ 
\hhline{|=::==::==|}
                       & \begin{tabular}[c]{@{}c@{}}Not soft\\ (Observed)\end{tabular} & \begin{tabular}[c]{@{}c@{}}Soft\\ (Observed)\end{tabular} & \begin{tabular}[c]{@{}c@{}}Not soft\\ (Observed)\end{tabular} & \begin{tabular}[c]{@{}c@{}}Soft\\ (Observed)\end{tabular}  \\
\hline
\end{tabular}%
}
\vspace{-1em}
\end{table}



\begin{figure*}[t]

     \centering
     \begin{subfigure}[b]{0.35\textwidth}
         \centering
         \includegraphics[width=\textwidth]{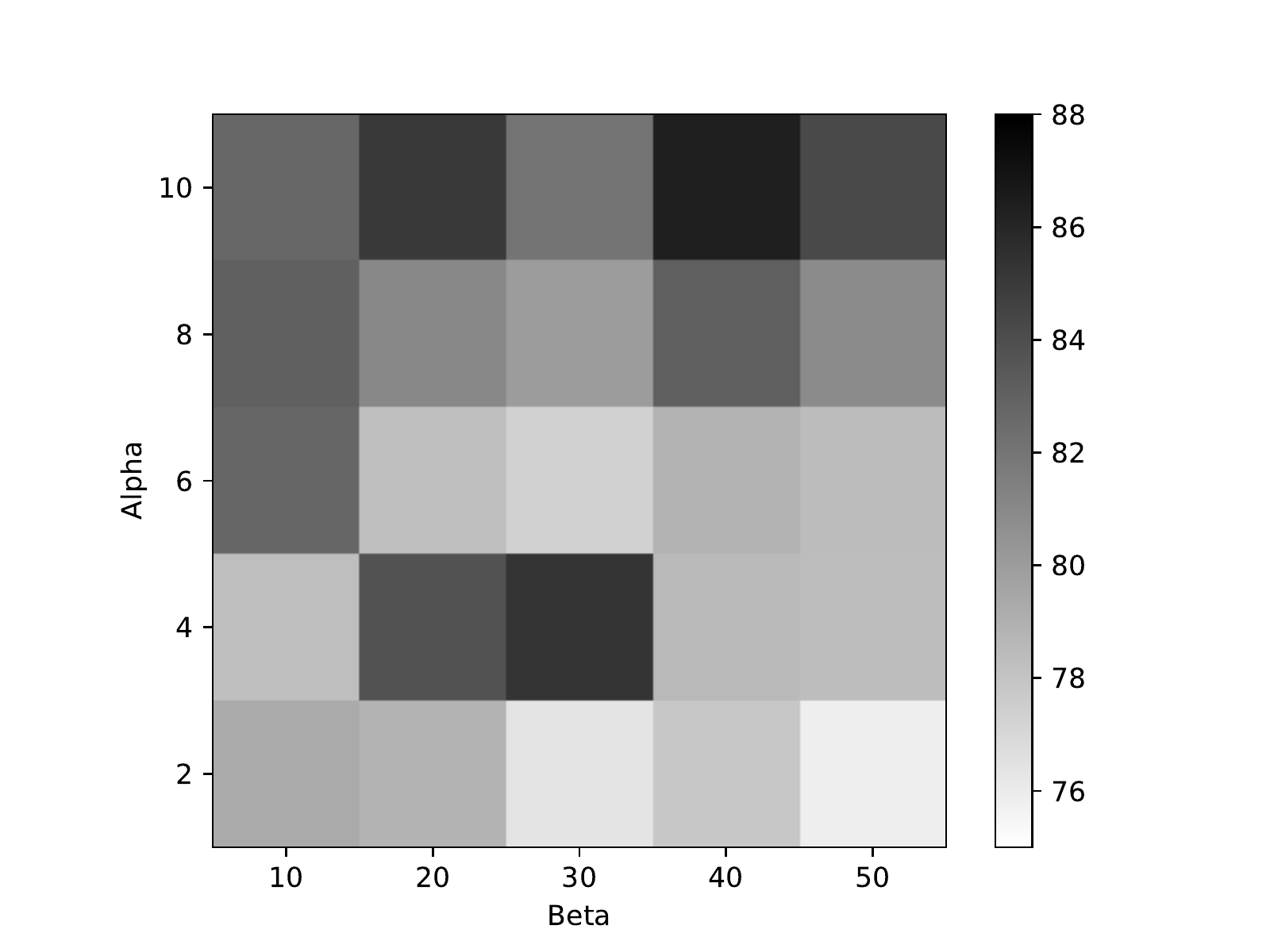}
         \vspace{-1.5em}
         \caption{Learning phase: early }
         \label{fig:param_early}
     \end{subfigure} \hspace{-5em}
     \begin{subfigure}[b]{0.35\textwidth}
         \centering
         \includegraphics[width=\textwidth]{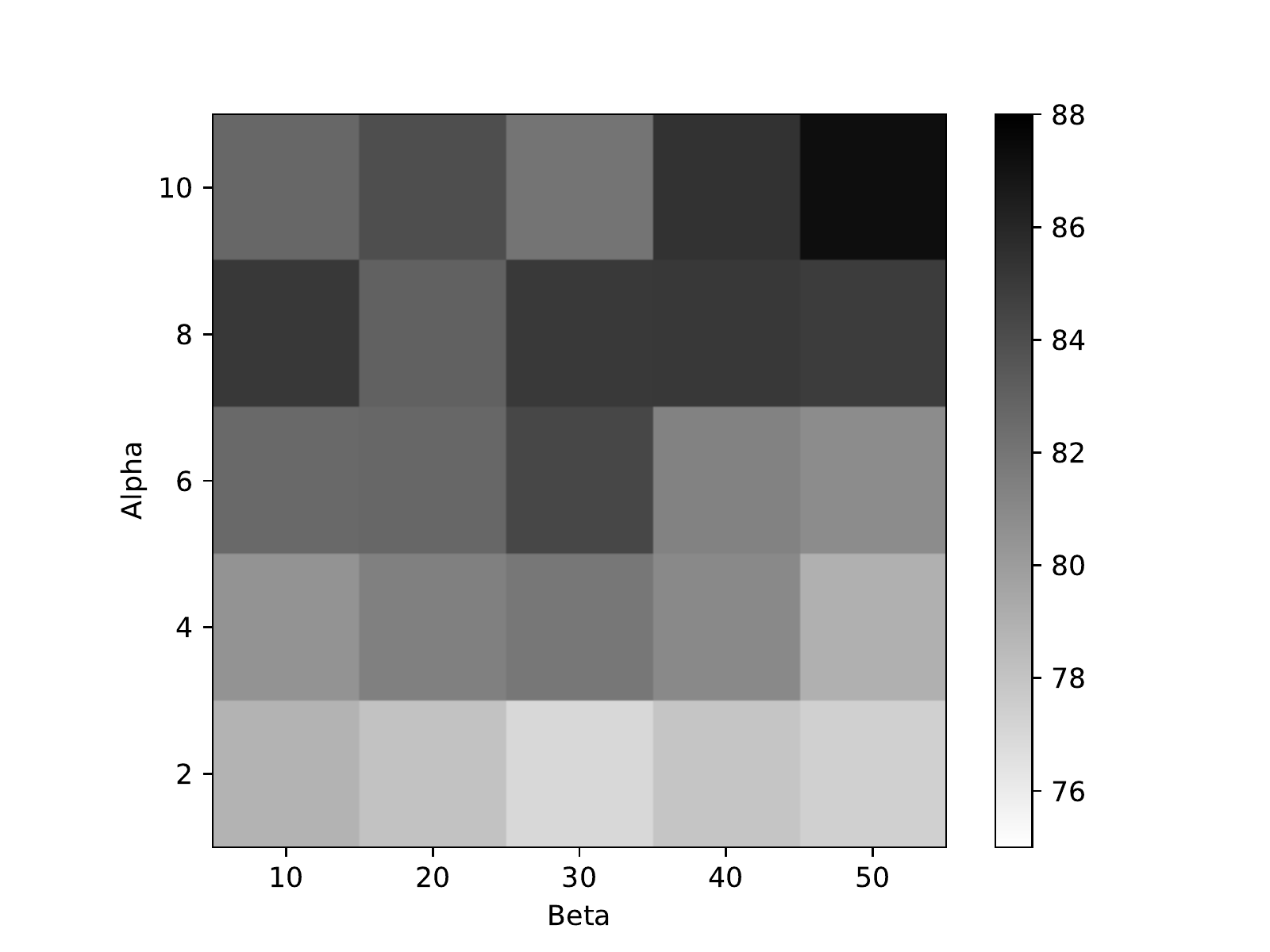}
         \vspace{-1.5em}
         \caption{Learning phase: middle }
         \label{fig:param_middle}
     \end{subfigure} \hspace{-5em}
     \begin{subfigure}[b]{0.35\textwidth}
         \centering
         \includegraphics[width=\textwidth]{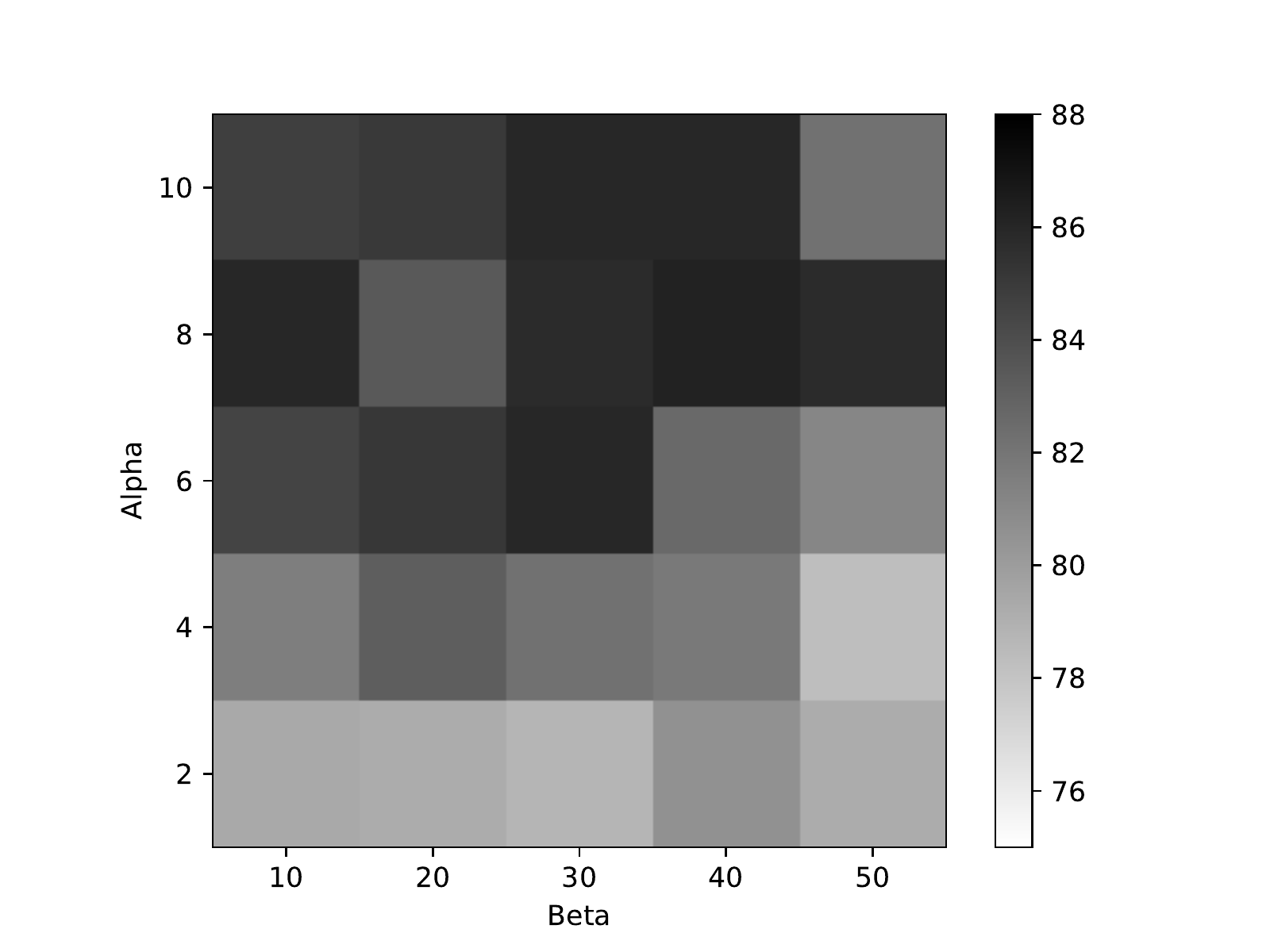}
         \vspace{-1.5em}
         \caption{Learning phase: late }
         \label{fig:param_late}
     \end{subfigure}
     \caption{We empirically evaluated the identification accuracies of ITRS using different values of $\alpha$ and $\beta$, which are two parameters of our reward shaping approach (Eqn.~\eqref{reward}). One observation is that the overall accuracy becomes higher in the middle and late learning phases, which is expected and verifies the robustness of ITRS to $\alpha$ and $\beta$ selections. Another observation is that the selections of $\alpha$ and $\beta$ affect the performance of ITRS. We leave the auto-learning of the parameters to future work.}
     \label{fig:param}
\end{figure*}

\vspace{.5em}
\subsection{Experimental Results}

\noindent
{\bf Exploration Cost and Success Rate: }
Figure~\ref{fig:pred} shows the learning curve for one-attribute and two-attribute identification query evaluated on \textbf{CY101} and \textbf{ISPY32}, where we conducted experiments over three different strategies (two baselines and ITRS).
$x$-axis is the accumulative cost over all trials at the training phase.
Since the cost is determined by the time required to complete each action, we can regard the $x$-axis as training time.
$y$-axis reflects the identification accuracy at the testing phase.
The proposed method consistently performs better in task-completion accuracy along the whole training process and achieves higher accuracy than baselines.

Although we provided the same pretraining data, three curves in the two subfigures (for each of the two datasets) do not start from the same point.
That is because pretraining data only affects the observation model for the robot, but it is not directly related to the policy for attribute identification.
Three strategies have the same observation model but they use different methods to select exploratory behaviors.
As a result, the task-completion accuracy is not the same for them at the starting point.
ITRS assigns extra rewards for exploration at the very beginning of the training phase.
It resulted in not only a bigger cost but also a higher accuracy.



\noindent
{\bf Individual Attributes: }
At an exploration cost budget of 2 hours, we further evaluated the performance of each individual attribute on \textbf{CY101} using the three strategies we mentioned, as shown in Figure~\ref{fig:ind}, where 10 attributes are ranked by the identification accuracy of our method, i.e. ITRS. 
The robot has a higher identification accuracy for most of the attributes using ITRS, while the random legal baseline produces a relatively weak result compared to the other two strategies.
Attributes such as ``plastic'', ``hard'', and ``empty'' are more difficult to learn since the accuracy is no more than 80\% for all three methods.
And attributes such as ``blue'', ``full'' and ``container'' are easier, where repeated assembly and ITRS both offer pretty good results.

\noindent
{\bf Parameters: }
In Eqn.~\eqref{reward}, we have two parameters $\alpha$ and $\beta$. 
We conducted experiments on \textbf{CY101} with different $\alpha$ and $\beta$ combinations, as shown in Figure~\ref{fig:param}.
Results show that overall, a small $\alpha$ value leads to a higher identification accuracy in the beginning, but the accuracy does not improve much when it reaches the middle or late learning phase.
A lower $\beta$ value encourages the robot to explore no matter whether it is experienced or not, while a higher $\beta$ value affects the robot to compute the optimal policy.
Thus, when $\beta$ is within the middle range, the robot has the best identification performance.

\begin{table}[t]
\centering
\caption{Actions, observations and belief updates in the demonstration trial.}
\label{tab:bf}
\resizebox{0.45\textwidth}{!}{%
\begin{tabular}{|c|c|c|c|}
\hline
Step & Action & Observation & \begin{tabular}[c]{@{}c@{}}Belief\\ (Initial belief: {[}0.5, 0.5{]})\end{tabular} \\ \hline
1    & Look   & Positive    & {[}0.41, 0.59{]}                                                                  \\ \hline
2    & Grasp  & Positive    & {[}0.33, 0.67{]}                                                                  \\ \hline
3    & Lift   & Positive    & {[}0.20, 0.80{]}                                                                  \\ \hline
4    & Shake  & Negative    & {[}0.83, 0.17{]}                                                                  \\ \hline
5    & Shake  & Positive    & {[}0.46, 0.54{]}                                                                  \\ \hline
6    & Shake  & Negative    & {[}0.94, 0.06{]}                                                                  \\ \hline
\end{tabular}%
}
\end{table}



\subsection{Real Robot Demonstration}
\label{sec:real}

We have demonstrated the learned action policy using a real robot (UR5e arm from Universal Robots). 
It should be noted that the two datasets we used in this research were collected on robots that are different from the robot in the demonstration. 
It is a major challenge in robotics of transferring skills learned from one robot to another. 
To alleviate the effect caused by the heterogeneity of robot platforms, after performing each action, we sampled a data instance from \textbf{CY101} according to $x \in \mathcal{X}$, the fully observable component of the current state. 

\begin{figure}[t]
     \begin{center}
     \begin{subfigure}[b]{0.32\columnwidth}
         \centering
         \includegraphics[width=\textwidth]{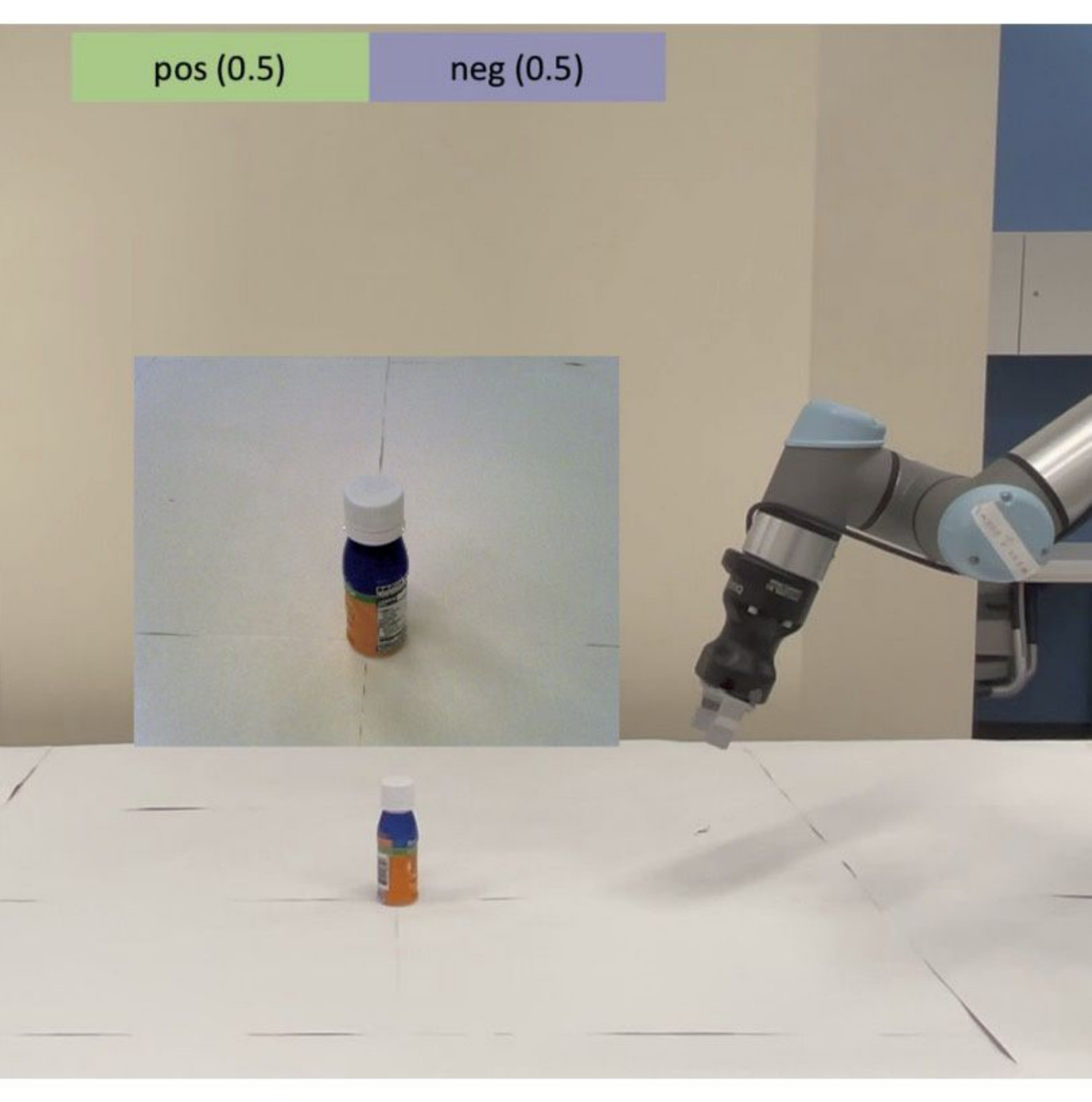}
         \vspace{-1.5em}
         \caption{$look$}
         \label{fig:look}
     \end{subfigure} 
     \begin{subfigure}[b]{0.32\columnwidth}
         \centering
         \includegraphics[width=\textwidth]{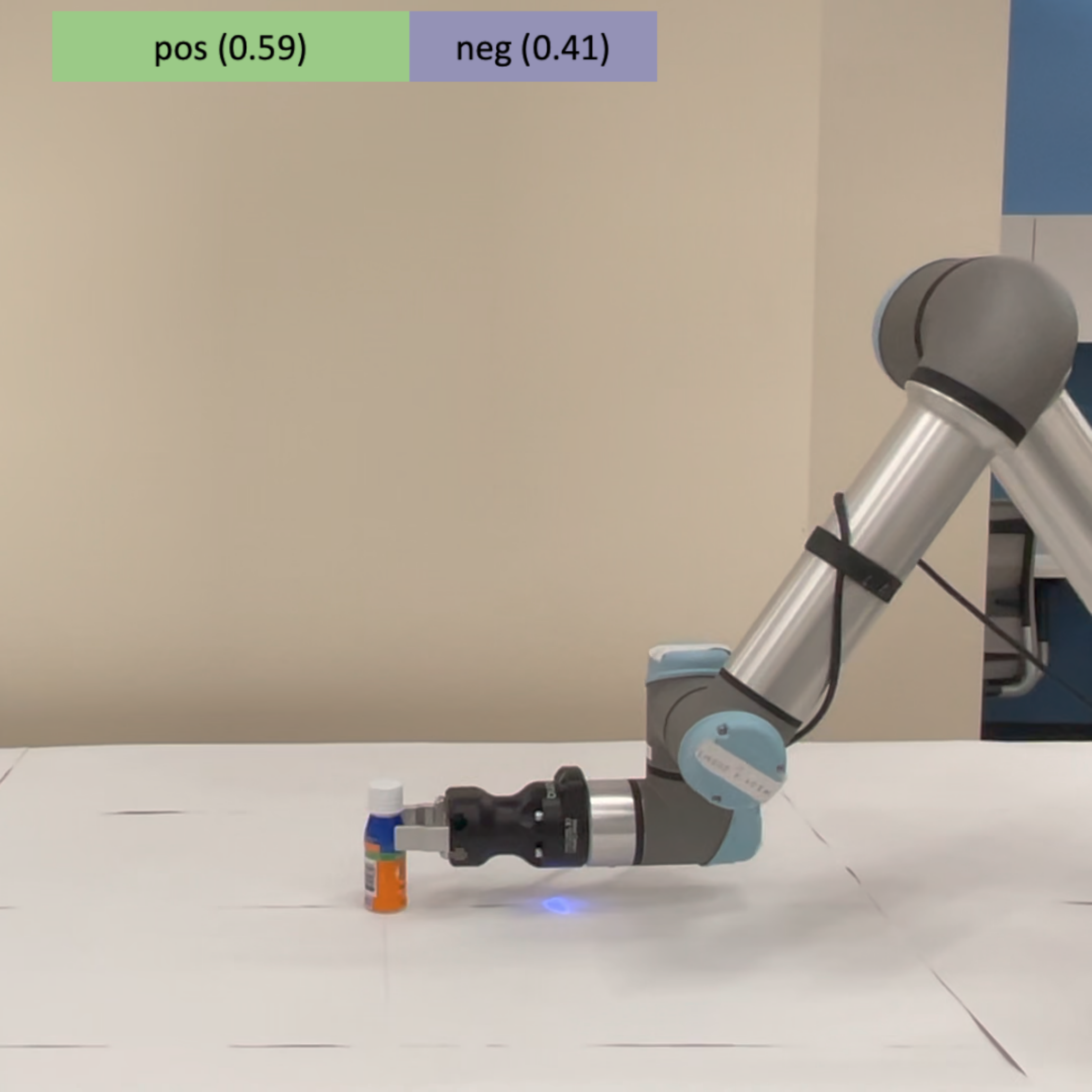}
         \vspace{-1.5em}
         \caption{$grasp$}
         \label{fig:grasp}
     \end{subfigure}
     \begin{subfigure}[b]{0.32\columnwidth}
         \centering
         \includegraphics[width=\textwidth]{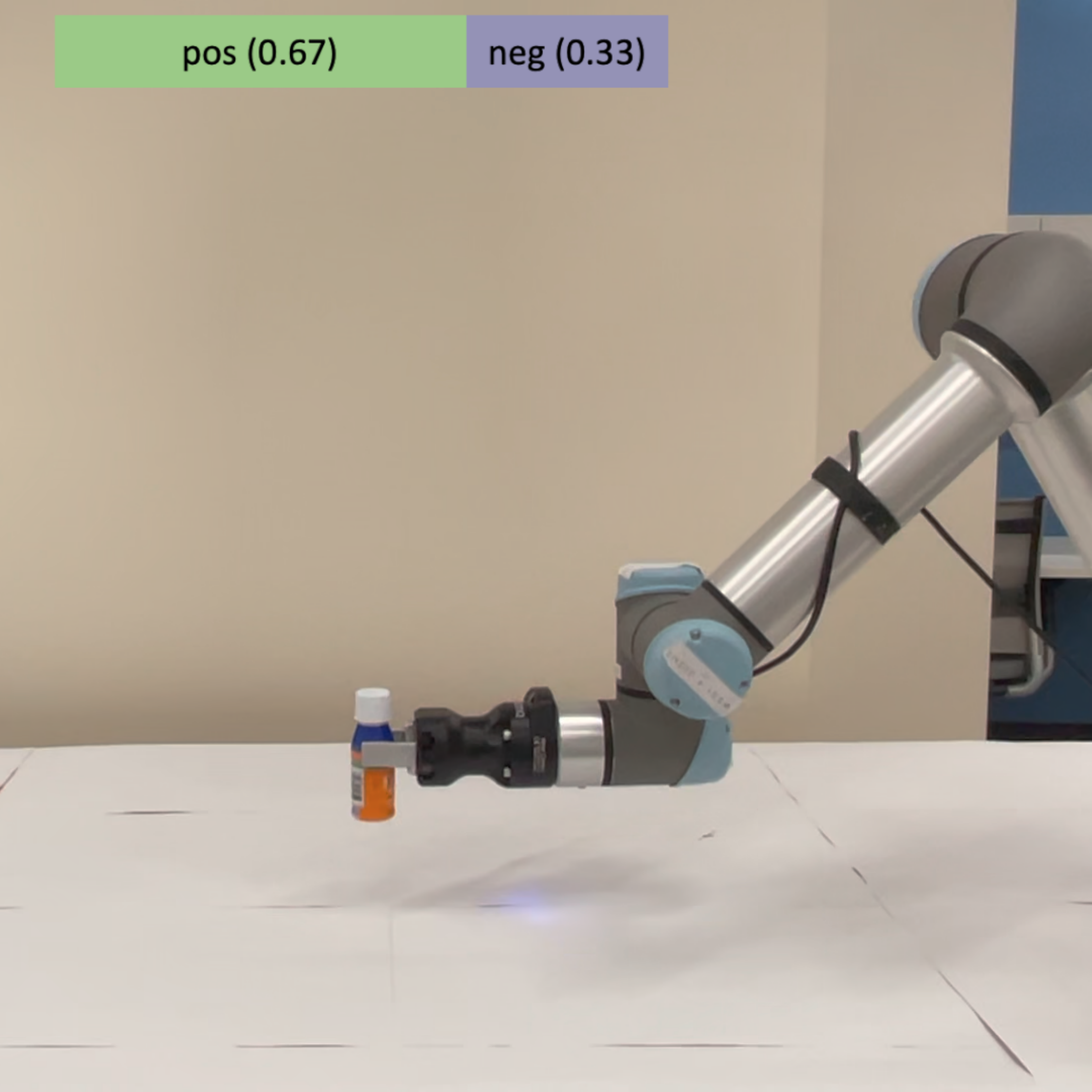}
         \vspace{-1.5em}
         \caption{$lift$}
         \label{fig:lift}
     \end{subfigure}
     \begin{subfigure}[b]{0.32\columnwidth}
         \centering
         \includegraphics[width=\textwidth]{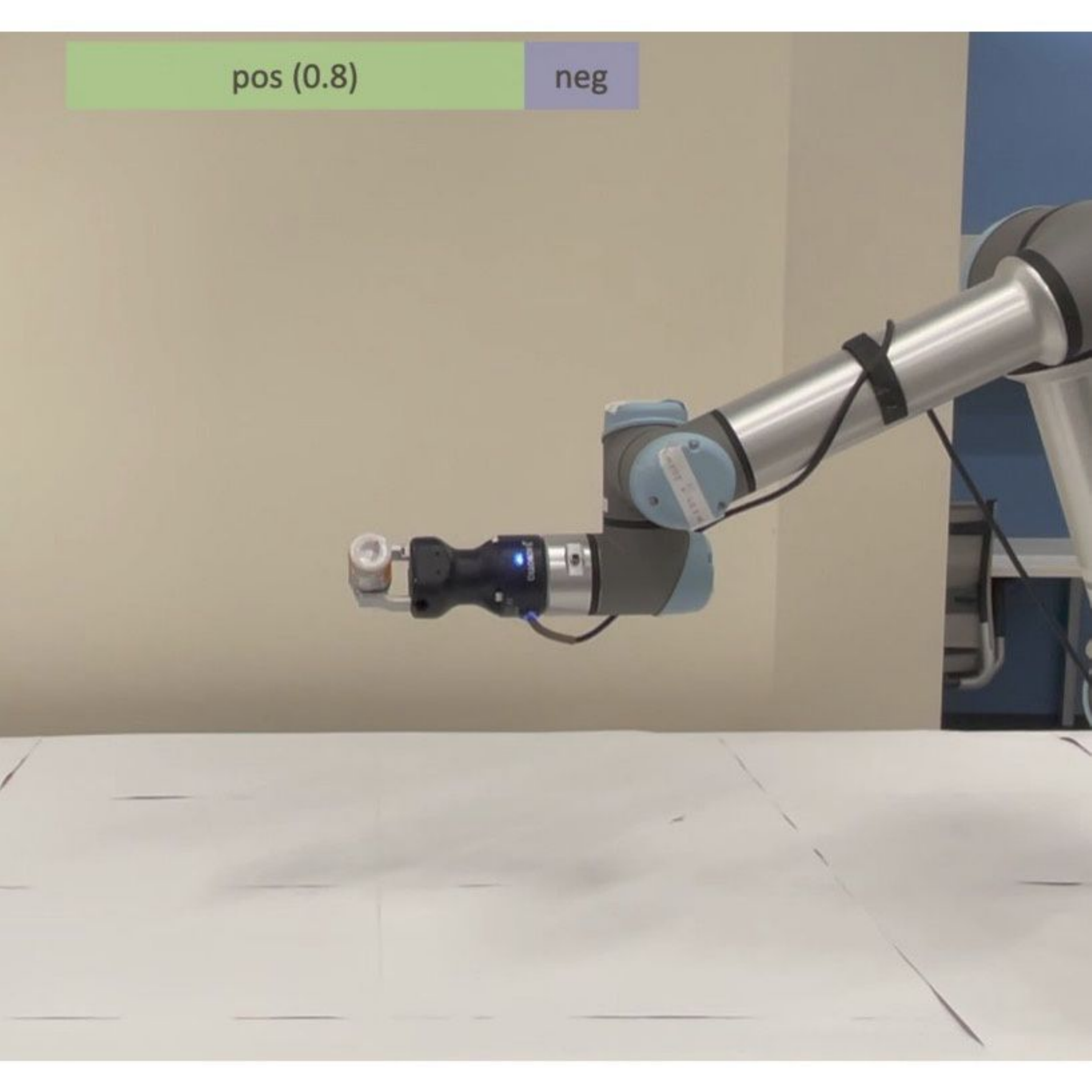}
         \vspace{-1.5em}
         \caption{$shake$}
         \label{fig:shake1}
     \end{subfigure}
     \begin{subfigure}[b]{0.32\columnwidth}
         \centering
         \includegraphics[width=\textwidth]{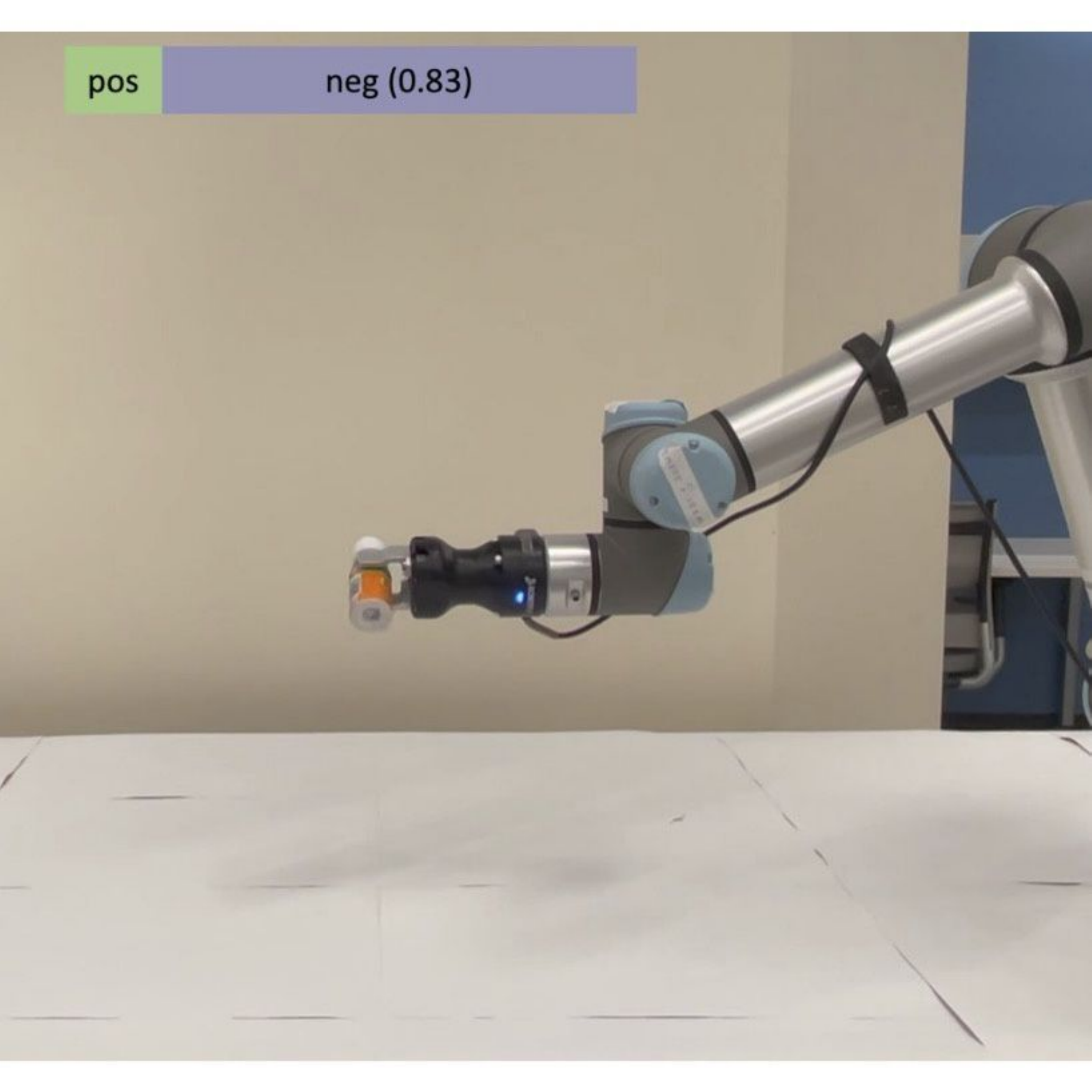}
         \vspace{-1.5em}
         \caption{$shake$}
         \label{fig:shake2}
     \end{subfigure}
    \begin{subfigure}[b]{0.32\columnwidth}
         \centering
         \includegraphics[width=\textwidth]{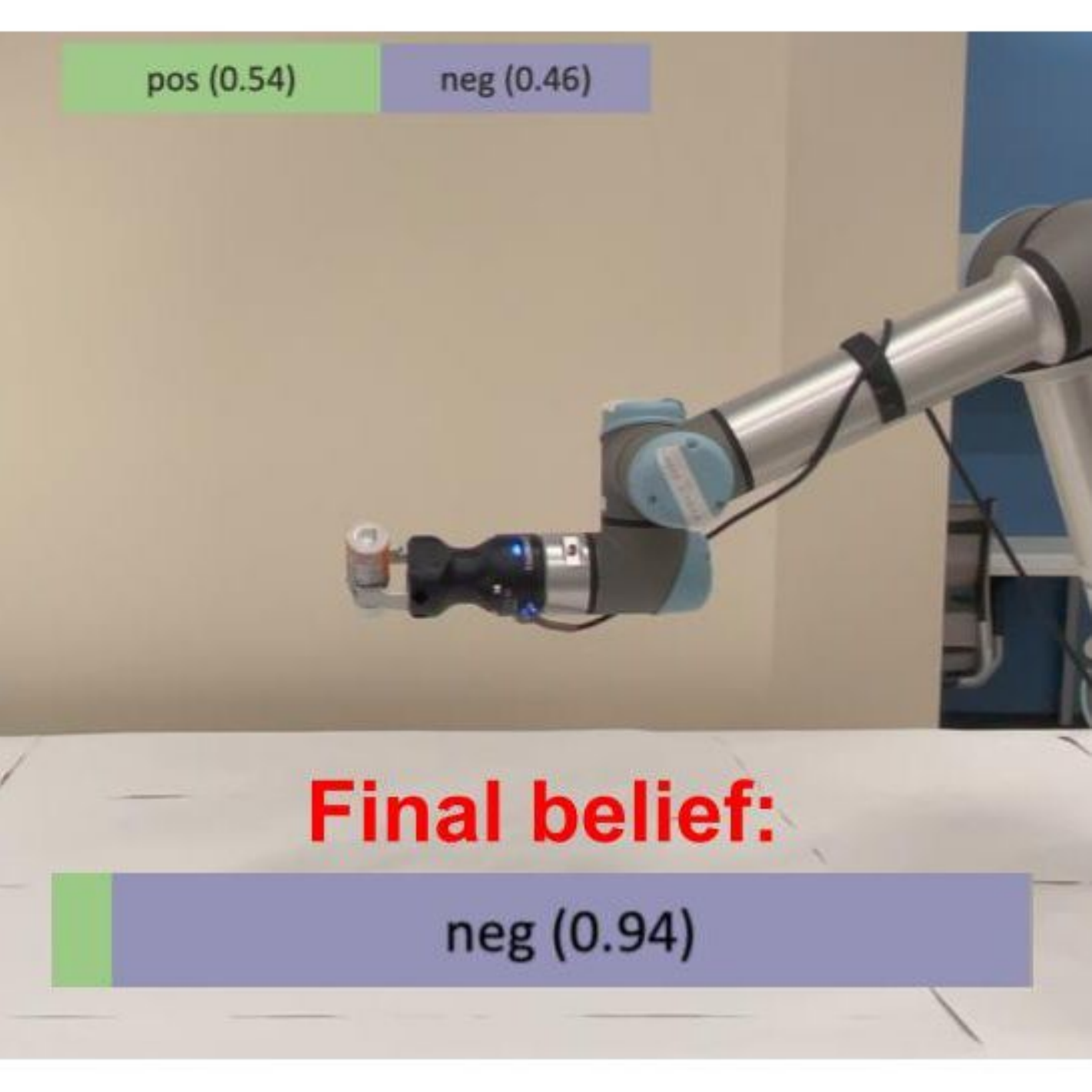}
         \vspace{-1.5em}
         \caption{$shake$}
         \label{fig:shake3}
     \end{subfigure}
     \end{center}     
     \vspace{-.5em}
     \caption{A demonstration of the learned action policy using ITRS algorithm. The robot performed six actions in a row. At the beginning, the robot started with a uniform distribution (it evenly believed the object can be empty or not). After completing the six actions, the belief converged to ``negative'' (0.94 probability). Finally the robot selected a reporting action to report that the object is ``\emph{not empty}.'' }
     \vspace{-1em}
     \label{fig:real}
\end{figure}

In the demonstration trial, our robot was given an object -- a pill bottle half-full of beans. 
The one-attribute query was ``\emph{Is this object empty?}''
The robot performed a sequence of exploratory actions, as shown in Table~\ref{tab:bf}, where we also listed the observation and the belief after each action. 
For instance, a ``Positive'' observation means that the robot perceives that the object is ``empty.''
Figure~\ref{fig:real} shows a sequence of screenshots of the UR5e robot completing the task using a learned ITRS policy. 

\section{Conclusions and Future Work}
In this work, we focus on a new On-RAL problem where the robot is required to complete attribute identification tasks and, at the same time, learn its observation model for each attribute.
We propose an algorithm called ITRS that selects exploratory actions toward simultaneous attribute learning and attribute identification.
The proposed method and baseline methods are evaluated using two real-world datasets. 
Experimental results show that ITRS enables the robot to complete attribute identification tasks at a higher accuracy using the same amount of training time compared to baselines. 




One limitation of our existing framework is that the attribute recognition models are learned by a single robot and cannot directly be used by another robot that has different behaviors, morphology, and sensory modalities. We plan to use sensorimotor transfer learning (e.g.,~\cite{tatiya2020haptic,Tatiya2020framework}) to scale up our framework to allow multiple different robots to learn such models and share their knowledge as to further speed up learning. 
In addition, considering correlations between attributes and handling fuzzy attributes can potentially improve the performance of On-RAL. 
Another direction for the future is to learn the world dynamics through the task completion process (currently the transition function is provided and the observation function is learned), where we can potentially use reinforcement learning methods. 
Finally, we plan to evaluate the learned policy using a robot platform that is different from the robot used for data collection, and with real human-robot dialogue to acquire attribute labels for objects.


\section*{Acknowledgments}
The authors thank Sinem Ozden, Yan Ding, Kishan Chandan, Kexin Li, and anonymous reviewers for helpful comments and discussions.
A portion of this research has taken place at the Autonomous Intelligent Robotics (AIR) Group, SUNY Binghamton. AIR research is supported in part by grants from the National Science Foundation (NRI-1925044), Ford Motor Company (URP Awards 2019-2021), OPPO (Faculty Research Award 2020), and SUNY Research Foundation.

\bibliographystyle{plainnat}

\bibliography{ref}

\end{document}